%% file: main.tex
\documentclass[10pt,twocolumn,letterpaper]{article}

\usepackage[pagenumbers]{cvpr} 

\input{preamble}
\usepackage{pifont}

\definecolor{cvprblue}{rgb}{0.21,0.49,0.74}
\usepackage[pagebackref,breaklinks,colorlinks,allcolors=cvprblue]{hyperref}

\def\confName{CVPR}
\def\confYear{2026}

\title{The Blind Spot of Adaptation: Quantifying and Mitigating Forgetting in Fine-tuned Driving Models}

\author{Runhao Mao$^{*}$, Hanshi Wang$^{*}$, Yixiang Yang, Qianli Ma, Jingmeng Zhou, Zhipeng Zhang$^{\dagger}$\\
AutoLab, School of Artificial Intelligence, Shanghai Jiao Tong University\\
{\tt\small amao769909148@gmail.com, zhipeng.zhang.cv@outlook.com}
}

\begin{document}
\maketitle

\begingroup
\renewcommand\thefootnote{}
\footnotetext{
$^{*}$ Equal contribution. $^{\dagger}$ Corresponding author.}
\endgroup
\input{sec/0_abstract}

\input{sec/1_intro}

\input{sec/2_Related_Works}

\input{sec/3_Fidelity_Bench}

\input{sec/4_Experiment}

\input{sec/5_Conclusion}
{
    \small
    \bibliographystyle{ieeenat_fullname}
    \bibliography{main}
}

\end{document}

%% file: preamble.tex
\newcommand{\TODO}[1]{\textbf{\color{red}[TODO: #1]}}
\usepackage{booktabs}   
\usepackage{multirow}   
\usepackage{makecell}
\usepackage{graphicx}
\usepackage{subcaption}
\usepackage[accsupp]{axessibility}

\graphicspath{{images/}}
  
\usepackage{siunitx}     
\renewcommand{\TODO}[1]{}

\usepackage{microtype}

\renewcommand{\paragraph}[1]{\vspace{.5em}\noindent\textbf{#1.}}

\usepackage{xspace}

\usepackage{soul}
\setuldepth{foobar}

\newcommand{\crbx}[2]{\scalebox{0.6}{\fcolorbox{#1}{#1}{\makebox[1.7ex][c]{\rule{0pt}{1.5ex}#2}}}}

\definecolor{crTSU}{RGB}{0, 150, 136}
\newcommand{\TSU}{\raisebox{+0.2ex}{\crbx{crTSU}{\textcolor{white}{\textbf{\textsf{U}}}}}}

\definecolor{crMA}{RGB}{0, 150, 136}
\newcommand{\MA}{\raisebox{+0.2ex}{\crbx{crMA}{\textcolor{white}{\textbf{\textsf{A}}}}}}

\definecolor{crTP}{RGB}{0, 150, 136}
\newcommand{\TP}{\raisebox{+0.2ex}{\crbx{crTP}{\textcolor{white}{\textbf{\textsf{TP}}}}}}

\definecolor{crDS}{RGB}{102, 187, 106}
\newcommand{\DSA}{\raisebox{+0.2ex}{\crbx{crDS}{\textcolor{white}{\textbf{\textsf{A}}}}}}  
\newcommand{\DSB}{\raisebox{+0.2ex}{\crbx{crDS}{\textcolor{white}{\textbf{\textsf{B}}}}}}  
\newcommand{\DSC}{\raisebox{+0.2ex}{\crbx{crDS}{\textcolor{white}{\textbf{\textsf{C}}}}}}  
\newcommand{\DSD}{\raisebox{+0.2ex}{\crbx{crDS}{\textcolor{white}{\textbf{\textsf{D}}}}}}  
\newcommand{\DSE}{\raisebox{+0.2ex}{\crbx{crDS}{\textcolor{white}{\textbf{\textsf{E}}}}}}  
\newcommand{\DSF}{\raisebox{+0.2ex}{\crbx{crDS}{\textcolor{white}{\textbf{\textsf{F}}}}}}  
\newcommand{\DSG}{\raisebox{+0.2ex}{\crbx{crDS}{\textcolor{white}{\textbf{\textsf{G}}}}}}  
\newcommand{\DSH}{\raisebox{+0.2ex}{\crbx{crDS}{\textcolor{white}{\textbf{\textsf{H}}}}}}  
\newcommand{\DSI}{\raisebox{+0.2ex}{\crbx{crDS}{\textcolor{white}{\textbf{\textsf{I}}}}}}  
\newcommand{\DSJ}{\raisebox{+0.2ex}{\crbx{crDS}{\textcolor{white}{\textbf{\textsf{J}}}}}}  
\newcommand{\DSK}{\raisebox{+0.2ex}{\crbx{crDS}{\textcolor{white}{\textbf{\textsf{K}}}}}}  
\newcommand{\DSL}{\raisebox{+0.2ex}{\crbx{crDS}{\textcolor{white}{\textbf{\textsf{L}}}}}}  
\newcommand{\DSM}{\raisebox{+0.2ex}{\crbx{crDS}{\textcolor{white}{\textbf{\textsf{M}}}}}}  
\newcommand{\DSN}{\raisebox{+0.2ex}{\crbx{crDS}{\textcolor{white}{\textbf{\textsf{N}}}}}}  
\newcommand{\DSO}{\raisebox{+0.2ex}{\crbx{crDS}{\textcolor{white}{\textbf{\textsf{O}}}}}}  
\newcommand{\DSP}{\raisebox{+0.2ex}{\crbx{crDS}{\textcolor{white}{\textbf{\textsf{P}}}}}}  
\usepackage{booktabs}  
\usepackage{xcolor}    
\usepackage{colortbl}  

%% file: sec/0_abstract.tex

\begin{abstract}
The integration of Vision-Language Models (VLMs) into autonomous driving promises to solve long-tail scenarios, but this paradigm faces the critical and unaddressed challenge of catastrophic forgetting. The very fine-tuning process used to adapt these models to driving-specific data simultaneously erodes their invaluable pre-trained world knowledge, creating a self-defeating paradox that undermines the core reason for their use. This paper provides the first systematic investigation into this phenomenon. We introduce a new large-scale dataset of 180K scenes, which enables the first-ever benchmark specifically designed to quantify catastrophic forgetting in autonomous driving. Our analysis reveals that existing methods suffer from significant knowledge degradation. To address this, we propose the Drive Expert Adapter (DEA), a novel framework that circumvents this trade-off by shifting adaptation from the weight space to the prompt space. DEA dynamically routes inference through different knowledge experts based on scene-specific cues, enhancing driving-task performance without corrupting the model's foundational parameters. Extensive experiments demonstrate that our approach not only achieves state-of-the-art results on driving tasks but also effectively mitigates catastrophic forgetting, preserving the essential generalization capabilities that make VLMs a transformative force for autonomous systems. Data and model are released at \href{https://github.com/AutoLab-SAI-SJTU/FidelityDrivingBench}{FidelityDrivingBench}.

\end{abstract}

%% file: sec/1_intro.tex
\section{Introduction}

\begin{figure}[!t]
  \centering
  \vspace{-0.5cm}
  \includegraphics[width=\linewidth]{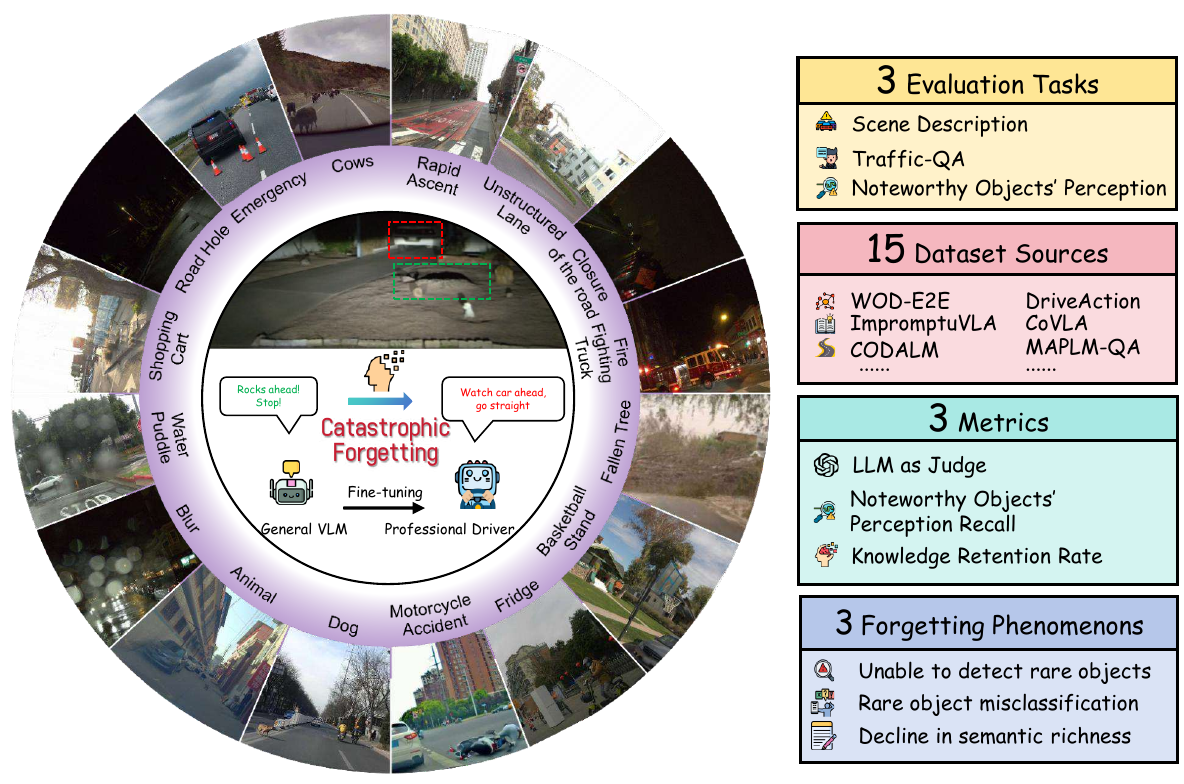}
  \vspace{-0.5cm}
  \caption{\textbf{Illustration of Fidelity Driving Bench}. We introduce a benchmark to quantify knowledge forgetting in general VLMs after fine-tuning on driving data, spanning 180K frames and 900K long-tail QA pairs, covering 3 tasks across 15 data sources with 2 forgetting metrics, and revealing 3 forgetting phenomena.}
  \label{fig:FidelityBench}
  \vspace{-0.5cm}
\end{figure}
\label{sec:intro}

Autonomous driving has undergone a significant shift from modular pipelines toward integrated end-to-end systems~\cite{planningorientedad,vad,e2eadgchallenges,trajectoryguidedcontrolpredictionendtoend,GE2EAD}. Nevertheless, these conventional end-to-end models struggle on long-tail scenarios due to limited world knowledge and commonsense for robust decisions~\cite{e2eadgchallenges}. Addressing these challenges, the advent of Vision-Language Models (VLMs)~\cite{llava, BLIP-2, Qwen2-VL, Flamingo, InternVL} has catalyzed a new VLM-centric paradigm for autonomous driving~\cite{drivegpt4, DriveMLM, LMDrive, Reason2Drive, AutoVLA}. This innovative approach utilizes language as a intermediate representation to unify perception, reasoning, and action within a cohesive system.

However, integrating VLMs into autonomous driving is not seamless due to a clear domain gap between general pretraining and driving scenes. To bridge this gap, contemporary methods fine-tune VLMs on specialized datasets, progressing from single dataset~\cite{DriveLM,BDD-X} to multi-source corpora such as ImpromptuVLA~\cite{impromptuvla}. Although these strategies yield impressive results on current benchmarks, they overlook a critical vulnerability. That is, any form of fine-tuning can induce catastrophic forgetting, which erodes the pre-trained world knowledge underlying generalization. This creates a fundamental paradox, since the very process used to adapt VLMs erodes their pre-trained world knowledge and in turn undermines the primary motivation for employing them to enhance generalization in unseen long-tail scenarios. 

\begin{figure}[t]
  \centering
  \includegraphics[width=\linewidth]{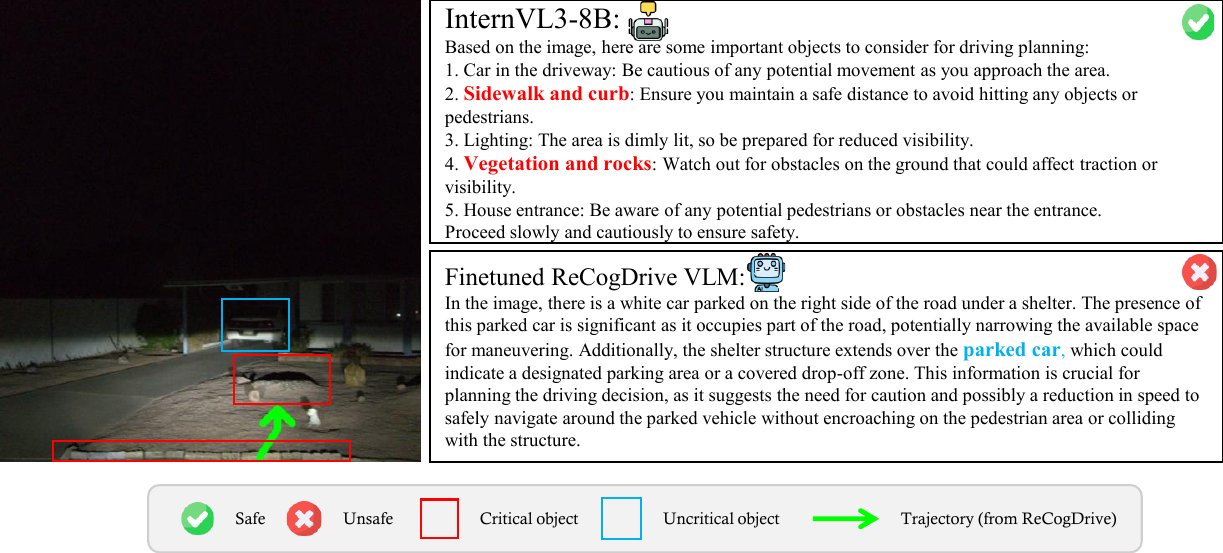}
  \caption{Catastrophic forgetting leads to degraded generalization in long-tail scenarios, which may result in safety-critical failures.}
  \label{fig:Catastrophic_forgetting_example}
  \vspace{-0.55cm}
\end{figure}

But strikingly, this is rarely explored in VLM-centric methods, and existing benchmarks are ill-equipped to even detect such degradation. To demonstrate the real-world consequences of this neglect, we conducted a visual analysis on long-tail scenarios. As shown in \cref{fig:Catastrophic_forgetting_example}, catastrophic forgetting causes RecogDrive~\cite{recogdrive} to overlook obstacles like curbs and rocks that its base model, InternVL3-8B~\cite{internvl3}, previously recognized, leading to unsafe trajectories. Our analysis further reveals this is a widespread issue (see appendix), posing a substantial safety risk. This brings us to the central question of our work: \textit{how can we quantitatively evaluate catastrophic forgetting in autonomous driving models?}

To answer the question, we realize that the most critical step is to construct a benchmark specifically engineered to measure this phenomenon. This is because existing benchmarks, even those focused on long-tail data, typically keep similar train and test distributions that obscure true knowledge loss. Therefore, we consolidate many datasets and enrich them by sampling and annotating corner cases with Qwen3VL-235B~\cite{Qwen-VL}. From this extensive data pool, we engineered a mining pipeline (see \cref{fig:datapipeline}) to systematically unearth rare scenes. Specifically, the pipeline extracts key elements from annotations (\eg, road conditions, traffic participants), assigns each an IDF-based (Inverse Document Frequency) rarity score, and calculates a scene's total rarity by summing the scores of its constituent elements. This method allowed us to filter 180K candidate scenes and, after manual inspection, distill a final forgetting test set of 1,000 images. The rest data forms what is, to our knowledge, the most comprehensive training set for VLM-centric driving models. Armed with this specialized benchmark, we then introduced the Knowledge Retention Rate (KRR) metric and evaluated a broad set of VLM-centric models. The results detailed in Sec.~\ref{analysis} revealed substantial forgetting across the board, validating our initial premise and underscoring the urgent need for effective mitigation strategies.

To investigate the root causes, we conducted a series of targeted experimental analyses. Our initial findings reveal that models fine-tuned on a single dataset exhibit significantly weaker performance and a lower KRR than those trained on matched-size multi-source data. This observation prompted a deeper analysis of existing benchmarks, where we identified the prevalent issue that they tend to focus on maximizing QA pair quantity at the expense of scene diversity (see \cref{tab:datasets comparison}). To address this gap, we constructed a new large-scale, diverse dataset of 180K scenes from 15 distinct sources for our subsequent experiments. Beyond dataset composition, we found the fine-tuning methodology itself is another critical factor. Our experiments show that full fine-tuning consistently leads to severe forgetting under domain shifts, while parameter-efficient methods like frozen layers or LoRA mitigate forgetting but yield suboptimal task performance (Tab.~\ref{tab:lora}). This inherent trade-off between knowledge retention and task-specific adaptation ultimately leads to the critical question of whether a method exists that can preserve prior knowledge while still effectively learning new tasks.

To address this critical trade-off, we contend that the solution lies not in altering the model's original parameters, but in intelligently selecting and activating the most relevant knowledge for a given task. While an intuitive first step might be to introduce new, trainable parameters via methods like LoRA, our experiments reveal this approach is insufficient. We found that LoRA struggles to fully bridge the domain gap between general pre-training data and specialized driving scenes, and it is susceptible to task-induced attention biases during fine-tuning. This motivated us to develop the \textbf{Driving Expert Adapter (DEA)}, a novel approach that shifts knowledge adaptation from the weight space to the prompt space. Specifically, our adapter dynamically routes inference through different driving experts based on scene-specific cues (\eg~, visibility, traffic density) and prompt semantics. As our experiments demonstrate, this method allows the model to maintain robust driving performance while effectively mitigating catastrophic forgetting.

Our main contributions are summarized as follows: \ding{171} We are the first to systematically investigate and identify the catastrophic forgetting in VLM-centric autonomous driving models.
\ding{170} We introduce a new large-scale dataset, which establishes the first systematic benchmark for forgetting evaluation and serves as the largest language-annotated driving corpus. \ding{168} We propose the Drive Expert Adapter, a novel framework that enhances adaptability to driving scenes while effectively mitigating forgetting by routing knowledge at the prompt level. \ding{169} Extensive experiments validate that our approach effectively mitigates catastrophic forgetting while achieving superior performance on driving-specific tasks.

%% file: sec/2_Related_Works.tex
\section{Related Works}
\label{sec:related_works}
\textbf{VLM-centric End-to-End Driving.} 
Recent end-to-end autonomous driving systems increasingly resort to VLMs to improve generalization. However, mainstream VLMs are designed to handle general tasks and cannot be directly applied to complex, high-real-time, and safety-demanding autonomous driving tasks. Most state-of-the-art VLM/VLA typically achieve superior performance by designing specialized architectures and tailored training strategies. For instance, some approaches incorporate 3D spatial information enhancement~\cite{opendrivevla, LMDrive, DriveMLM, Reason2Drive, OmniDrive} and temporal information modeling~\cite{sce2drivex, futuresightdrive, hilm-d} to better capture the dynamics of driving environments. Some research explores various data-efficient learning strategies such as knowledge distillation~\cite{fed, dsdrive}, reinforcement learning~\cite{alphadrive, AutoVLA, recogdrive, autodrive-r2}, and self-supervised learning~\cite{reasonplan, dilu}.
To enable more effective fine tuning, many related studies have also constructed large scale driving VQA datasets. DriveLM \cite{DriveLM} introduces a Graph-VQA task by creating annotations in the nuScenes \cite{nuScenes} and CARLA \cite{Carla} datasets. LingoQA \cite{lingoqa} and CoVLA~\cite{covla} collect a large number of video clips to generate extensive QA pairs. The CODA-LM Dataset \cite{codalm} specifically targets corner cases in autonomous driving scenarios. ImpromptuVLA \cite{impromptuvla} constructs a multi-source dataset focusing on unstructured corner case scenes.  OmniDrive \cite{OmniDrive} leverages the 3D spatial information provided by nuScenes to generate QA pairs~ aiming to enhance the spatial perception ability of VLMs. Although existing methods perform well on current benchmarks, fine tuning inevitably induces catastrophic forgetting and erodes the broad generalization capabilities that motivate the use of VLMs in autonomous driving. However, none of these efforts consider the forgetting of VLMs during fine tuning, and there is no dedicated dataset for evaluation.

\noindent\textbf{Catastrophic Forgetting in Post-training of VLMs.}
While effective post-training strategies can substantially enhance a model’s final performance on specific tasks, they also lead to catastrophic forgetting. To address this issue, prior work mainly explores three types of solutions, including architecture-based approaches~\cite{continual-llava, moelora, llava-cmoe, modalprompt}, regularization-based approaches~\cite{moincl, llava-c}, and replay-based approaches~\cite{vqacl, protogroup}. For instance, EProj~\cite{EProj} alleviates catastrophic forgetting by dynamically selecting Q-Former layers based on multimodal similarity, coupled with adaptive regularization and replay of previously seen samples. MoELoRA \cite{moelora} incorporates LoRA modules as expert components in a mixture-of-experts architecture, where a learnable router enables dynamic expert selection. In this paper, we systematically analyze forgetting induced by fine tuning in autonomous driving scenarios, establish the first benchmark for this problem. Based on extensive experiments on our benchmark, we further propose the Driving Expert Adapter to effectively mitigate knowledge degradation.

%% file: sec/3_Fidelity_Bench.tex
\begin{figure*}[!htbp]
  \centering
    \includegraphics[width=\linewidth]{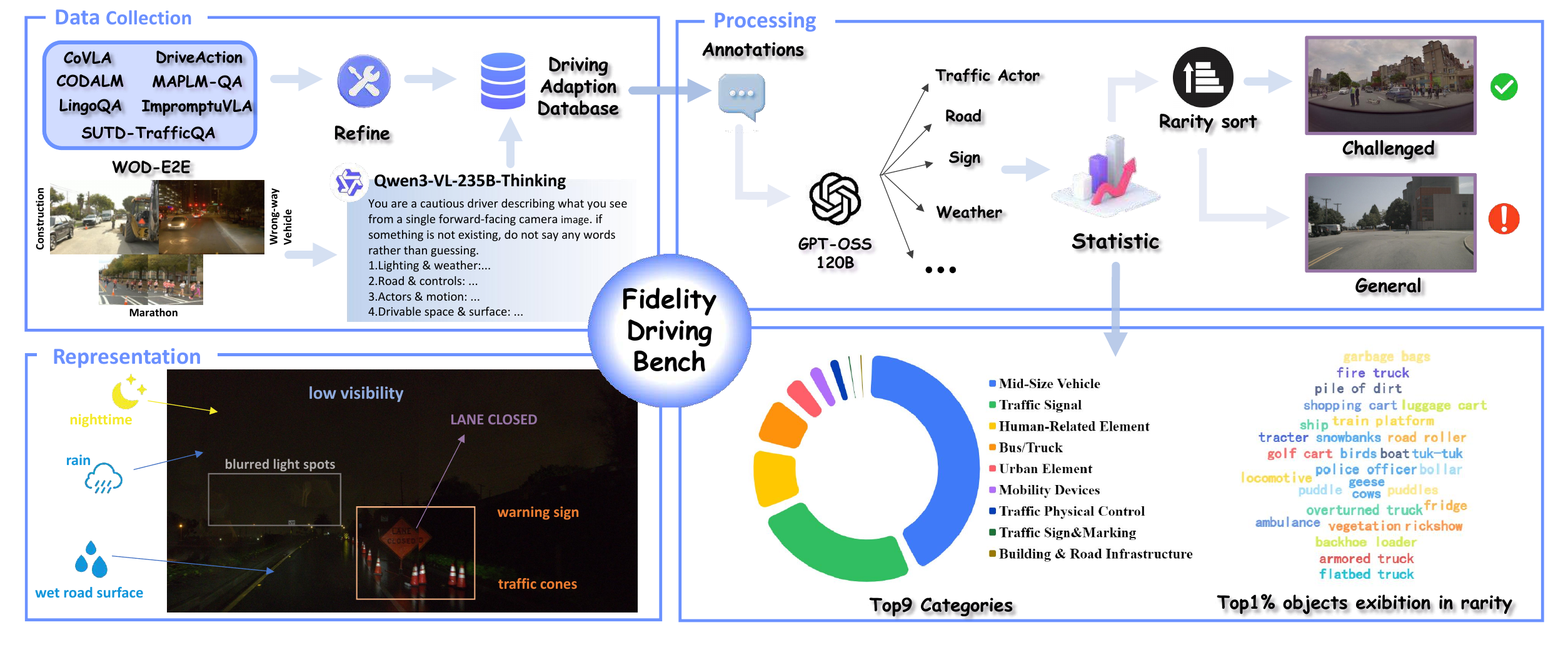}
    \caption{\textbf{The proposed dataset construction pipeline.} We first integrated fifteen existing annotated datasets and additionally provided language annotations for the WOD-E2E dataset. After that, each scene is represented by a set of sparse elements, which are automatically extracted from annotations using GPT. Then, we conduct manual verification to ensure accuracy and diversity. Finally, we retain 1,000 representative images as the test set, and the corresponding statistical summaries are presented in the bottom-right pane.}
    \label{fig:datapipeline}
    \vspace{-0.5cm}
\end{figure*}
\vspace{-0.1cm}
\section{Fidelity Driving Bench}
\vspace{-0.05cm}
\subsection{Benchmark and Dataset}
\vspace{-0.05cm}
\label{sec_31}
Although human drivers can recognize rare scenarios, scaling a dataset through manual curation alone is impractical. We therefore design an automated pipeline to acquire long-tail scenarios at scale. As illustrated in \cref{fig:datapipeline}, the mining process expands scene coverage far beyond what manual efforts can achieve. Specifically, we leverage GPT-OSS-120B~\cite{openai2025gptoss120bgptoss20bmodel} to extract meaningful scene elements from language annotations, compute their statistical frequency across datasets, and assign each element an IDF(Inverse Document Frequency) rarity score to identify long-tail scenarios. After ranking and manual verification, we selected 1,000 representative scenes for evaluating VLM forgetting in autonomous driving.
\begin{table}[t]
\caption{The Comparison of autonomous driving vision-language datasets. VL task icons: \TSU~Scene Understanding, \MA~Motion Analysis, \TP~Trajectory Planning.}
\label{tab:datasets comparison}
\centering
\resizebox{\columnwidth}{!}{
\begin{tabular}{lcccccc}
\toprule
\textbf{Dataset} & Frame & Video & QA Pairs & VL Tasks & Sources \\
\midrule
DriveLM-nuScenes~\cite{DriveLM} & 4.8K & - & 443K & \TSU~\MA~\TP & 1 \\
MAPLM-QA~\cite{maplm} & 14K & - & 61K & \TSU & 1  \\
CODALM~\cite{codalm} & 10K & - & 21K & \TSU~\MA & 1  \\
DriveBench~\cite{drivebench} & 19K & - & 20K & \TSU~\MA & 1  \\
CoVLA~\cite{covla} & - & 10K & 6M & \TSU~\MA & 1  \\
OmniDrive~\cite{OmniDrive} & 34K & - & 544K & \TSU~\MA~\TP & 1  \\
LingoQA~\cite{lingoqa} & 28K & - & 419K & \TSU~\MA & 1  \\
ImpromptuVLA~\cite{impromptuvla} & 80K & - & 560K & \TSU~\MA~\TP & 8  \\
VLADBench~\cite{vladbench} & 2K & 3K & 12K & \TSU~\MA & 12  \\
\midrule
\rowcolor{blue!8}
\textbf{Fidelity Driving Bench} & \textbf{180K} & - & 900K & \TSU~\MA & \textbf{15} \\
\bottomrule
\end{tabular}
}
\vspace{-0.5cm}
\end{table}

\subsubsection{Automatic Scene and Object Statistics}
Building on the forgetting phenomena summarized in Sec.~\ref{sec_31}, we identify long-tail scenarios in autonomous driving by conducting a frequency-based statistical analysis across the merged datasets. 

\noindent \textbf{Consolidating Dataset with Language Annotation.} First, we collected 14 most commonly used autonomous driving datasets that are annotated with language descriptions, including but not limited to DriveLM \cite{DriveLM}, ImpromptuVLA \cite{impromptuvla}, and CODALM \cite{codalm}, and unified them into a question–answer (QA) format, producing a million-level dataset that covers diverse driving scenes, object categories, and reasoning types. Next, we built a traffic-scene statistics pipeline based on GPT-OSS-120B to automatically analyze and categorize annotation content. Each annotation was classified into \textit{Traffic Scene Understanding} and \textit{Meta Action Decision} according to its semantics. We then focused on analyzing and summarizing the scene types and object categories appearing within the Traffic Scene Understanding subset. Objects in autonomous driving scenes are grouped into five major and fourteen minor categories. The bottom right of \cref{fig:datapipeline} presents the nine most frequent sub-categories, ordered from high to low frequency, including \textit{Mid-Size Vehicle, Traffic Signal, Human-Related Element, Bus/Truck, Urban Element, Mobility Devices, Traffic Physical Control, Traffic Sign \& Marking, and Building \& Road Infrastructure}. The remaining five sub-categories with the lowest occurrence frequencies, in descending order, are \textit{Natural Elements, Industrial Vehicles, Heteromorphic Vehicles, Specialized Vehicles, and Animals}. After consolidating, merging, and filtering all instances, \cref{fig:datapipeline} visualizes a dozen representative long-tail terms with extremely low occurrence frequencies across the five major categories, \textit{e.g., police officer, ship, geese}.

\noindent \textbf{Data Expansion.} Despite our integration of multiple datasets, the resulting collection still offered an insufficient volume of long-tail data. To further address the scarcity of these scenarios, we turned to the recently published WOD-E2E~\cite{wode2e}. Its diverse and large-scale nature provides a richer source of the very long-tail scenes that are underrepresented in other datasets, making it an ideal candidate for augmenting our evaluation set. We therefore leverage the powerful Qwen3VL-235B model~\cite{Qwen-VL} to add a rich semantic layer to this dataset, enriching each sample with a comprehensive scene description. Finally, these annotations are unified with the pre-existing datasets to form a more comprehensive and reliable foundation for subsequent long-tail scene mining.


\subsubsection{Frequency Guided Long Tail Mining}
\label{sec:datamining}

\noindent\textbf{Long-tail Scenarios Mining.} 
To obtain more reliable statistics, we transform each scene into a quantifiable representation. We develop a sparse scene representation that transforms each scene to a set of scenario elements $\{e_1,e_2,\dotsc,e_n\}$. Then we calculate per-element frequencies from statistical result and assign an IDF rarity~\cite{IDF}: 
\begin{equation} r(e) = \log\frac{N+\alpha}{n_e+\alpha},\label{eq:rarity}\end{equation}
where \(N\) denotes total number of images, \(n_e\) is number of images that contain element \(e\), and \(\alpha>0\) is a Laplace-smoothing constant that avoids extreme cases (\eg, \(n_e=0\)) and mitigates the overemphasis of ultra-rare terms.



To compute the rarity score \(R\) of each image, we sum the per-element rarity values within the image and then apply a BM25-style \emph{length normalization} so that images with long object lists do not receive unfairly high scores merely by enumerating many common objects. 
\begin{equation}
R(i)
= \Bigg(\sum \mathrm{r}(e)\Bigg)
  \times
  \underbrace{\frac{k+1}{k\,(1-b + b\,n_{o}^{i}/n_{o}^{avg}) + 1}}_{\text{BM25 length normalization}},
\end{equation}
where \(n_{o}^{i}\) denotes the number of objects in the image \(i\), and \(n_{o}^{avg}\) is the average number of objects per image in dataset. \(k\in[1.2,2.0]\) and \(b\in[0,1]\) control the strength of the length normalization.
Intuitively, if two images contain the same rare objects but one also includes many frequent objects, the length normalization will reduce the \(R\) of image, preventing inflated scores due to sheer quantity. \\
\textbf{Human Review and Annotation.} In the previous step, we selected approximately 2,500 images corresponding to the top 1.25\% rarity percentile scenes. A primary challenge was the inconsistent annotation formats and quality, as these images were sourced from datasets originally annotated by different models. To establish a unified, high-quality standard for evaluation, we first employed Qwen3VL-235B to re-annotate the entire pool. Using carefully designed prompt templates, this process generated consistent and highly detailed \textit{Scene Descriptions} and identified \textit{Noteworthy Objects}. Subsequently, the entire auto-annotated set underwent a rigorous manual verification. This not only filtered the collection down to the 1,000 most challenging and uncommon scenes but also served to meticulously correct any inaccuracies and completely eliminate model-induced hallucinations, thereby guaranteeing the final test set's fidelity.

\noindent\textbf{Training Set Construction.} Beyond the curated test set, we processed the vast remainder of our initially collected about 200K scenes to construct a large-scale, high-quality training dataset. Following a rigorous refinement protocol, we systematically removed low-quality image-annotation pairs, standardized multi-view datasets by retaining only front-view images for consistency, and distilled sequential clips into single, representative key frames, pruning any QA pairs rendered obsolete by this consolidation. The culmination of this effort is a massive, unified training dataset of 180K driving scenes from 15 distinct sources, richly annotated with over 900K QA pairs. To our knowledge, this represents the most extensive language-annotated training corpus for autonomous driving to date.



\begin{table*}
\caption{Evaluation of VLM-centric autonomous driving models on our proposed benchmark. Dataset icons: \DSA~DriveLM, \DSB~LingoQA, \DSC~DRAMA, \DSD~Talk2Car, \DSE~SUTD, \DSF~NuScenesQA, \DSG~DriveGPT4, \DSH~MAPLM, \DSI~NuInstruct, \DSJ~CODALM, \DSK~OmniDrive, \DSL~Senna, \DSM~NAVSIM, \DSN~BDDX, \DSO~Impromptu VLA, \DSP~LMDrive.}
\label{tab:advlm evaluation}
\centering
\resizebox{0.98\textwidth}{!}{
\begin{tabular}{lccccccc}
\toprule
\textbf{Method} & \textbf{Source} & \textbf{Datasets} & \textbf{Base} & \textbf{SD} $\uparrow$ & \textbf{T-QA} $\uparrow$ & \textbf{NoPR} $\uparrow$ & \textbf{KRR} $\uparrow$ \\
\midrule
RecogDrive-8B~\cite{recogdrive}  & ICLR-2026
&\DSA\,\DSB\,\DSC\,\DSD\,\DSE\,\DSF\,\DSG\,\DSH\,\DSI\,\DSJ\,\DSK\,\DSL\,\DSM & InternVL3-8B  &65 & 45.5 & 41.3 & 74\%\\
RecogDrive-2B~\cite{recogdrive} & ICLR-2026 & \DSA\,\DSB\,\DSC\,\DSD\,\DSE\,\DSF\,\DSG\,\DSH\,\DSI\,\DSJ\,\DSK\,\DSL\,\DSM & InternVL3-2B& 63.7 & 42.2 & 36.1 & 71\% \\
RoboTron-Drive~\cite{robotron-drive}& ICCV-2025 & \DSJ\,\DSH\,\DSA\,\DSB\,\DSK\,\DSI & LLaMA-3.1+Siglip & 61.9 & 28.2 & 30.7 & -  \\
safeAuto-BDDX~\cite{safeauto} & ICML-2025 & \DSN & Video-LLaVA & 25.2 & 20.1 & 6.1 & 31.5\%\\
safeAuto-DriveLM~\cite{safeauto} & ICML-2025 & \DSA & Video-LLaVA & 27.4 & 10.2 & 8.3 & 41.8\%\\
ImpromptuVLA-7B~\cite{impromptuvla} & ICCV-2025 & \DSO & QwenVL2.5-7B & 60.8 & 46.3 & 32.2 & 67.5\%  \\
ImpromptuVLA-3B~\cite{impromptuvla}& ICCV-2025 & \DSO & QwenVL2.5-3B & 59.1 & 33.0 & 25.2 & 68.4\% \\
WiseAD~\cite{WiseAD}& arXiv & \DSB\,\DSC\,\DSP & MobileVLM-V2-1.7B & 47.8 & 19.8 & 5.2 & 30\% \\
Mini-Intern-DriveLM~\cite{mini-internvl} & CVPR 2024 challenge & \DSA & InternVL2-4B & 39.9 & 33.1 & 13.3 & 35.9\% \\
Mini-Intern-BDDX~\cite{mini-internvl} &CVPR 2024 challeng  & \DSN & InternVL2-4B & 51.1 & 33.8 & 19.8 & 53.5\% \\
\bottomrule
\end{tabular}
}
\vspace{-0.5cm}
\end{table*}



\begin{table}
\caption{Evaluation of different foundation VLMs (without finetuning) on our proposed benchmark.}
\label{tab:AD model comparison}
\centering
\resizebox{0.95\columnwidth}{!}{
\begin{tabular}{lcccccc}
\toprule
\textbf{Name}  & \textbf{SD} $\uparrow$ & \textbf{T-QA} $\uparrow$ & \textbf{NoPR} $\uparrow$  \\
\midrule
InternVL3-8B & 66.8 & 42.4 & 55.2 \\
InternVL3-2B & 63.8 & 31.5 & 50.8 \\
InternVL2-1B & 57.5 & 20.9 & 35.6 &  \\
InternVL2-4B &  61.3 & 30.0 & 37.1 \\
MobileVLM-V2-1.7B & 48.1 & 19.5 & 19.7 \\
QwenVL2.5-3B &  56.6 & 28.7 & 36.8  \\
QwenVL2.5-7B &  65.1 & 31.0 & 47.7  \\
Video-LLaVA &  50.1 & 25.2 & 19.1 \\
LLAVA-OV-0.5B &  56.8 & 33.7 & 26.2 \\
LLAVA-OV-7B &  63.7 & 35.9 & 43.3 \\
\bottomrule
\end{tabular}
}
\vspace{-0.5cm}
\end{table}




\subsection{Evaluation Tasks and Metrics}
\label{sec:eval}
\subsubsection{Tasks}
\noindent\textbf{Scene Description.}~This task aims to evaluate the model’s visual grounding and attentional alignment in autonomous driving scenes. Without any handcrafted or guiding prompts, the model is instructed to freely perceive and describe all observable information within the scene. 

\noindent\textbf{Traffic-QA.}~This task evaluates the model’s traffic scene comprehension and decision-making abilities by posing questions related to driving decisions and scenarios encountered in traffic environments.

\noindent\textbf{Noteworthy Objects' Perception.}~Without any handcrafted or guiding prompts, this task requires the model to perceive key objects in the scene to the best of its ability.

\subsubsection{Metrics}
\label{sec:metrics}

We propose a suite of complementary metrics that assess models from multiple perspectives. These metrics measure autonomous driving competence and quantify forgetting.

\noindent\textbf{LLM as Judge.} While conventional NLP metrics such as BLEU \cite{bleu}, METEOR \cite{meteor}, and CIDEr \cite{cider} measure lexical overlaps between generated and reference texts, they are inherently limited in evaluating semantic fidelity and safety-critical reasoning. In autonomous driving, the correctness of a response often depends on contextual understanding (\textit{e.g.}, whether the model recognizes a red traffic light or yields to pedestrians) rather than surface-level wording. Such nuances are beyond the capability of traditional n-gram–based metrics. In recent works, LLM-based evaluators have shown a more robust and interpretable alternative. They possess strong semantic comprehension and reasoning abilities, allowing them to assess task-specific correctness under natural language rubrics. In our Fidelity Driving Bench, we employ different LLMs (\textit{i.e.,} Gemini2.5-Flash, GPT-4o, Qwen3-Max) as a unified evaluator across all three tasks of scene description, traffic-qa, and noteworthy objects' perception. Each evaluation is guided by a task-specific prompt template and structured scoring criteria. This design ensures that evaluations reflect both linguistic coherence and domain-specific safety requirements. In the subsequent experiments, we present only the results from Qwen-judge, given that the test data was also annotated using Qwen-VL. A detailed analysis of different LLM-judges in detailed in Sec.~\ref{sec:llm_judge_ex}. 
The evaluation prompt templates, is provided in the Appendix.

\noindent\textbf{Noteworthy Objects' Perception Recall. }To rigorously assess model forgetting in long-tail scenarios, we introduce the criterion of object-level perception. As aforementioned, our long-tail scenes are inherently populated with rare objects (see \cref{fig:datapipeline} again). A model’s failure to identify these objects serves as a high-fidelity litmus test for pinpointing knowledge erosion after fine-tuning. And most importantly, the stakes of perceptual failure are highest at the object level. An inability to see a tangible obstacle, such as an animal or debris, presents a direct and catastrophic collision hazard.

Guided by this principle, we developed a quantitative protocol. Specifically, the LLM judge evaluates the model's textual output against these ground-truth annotations, assigning one of three labels: ``clearly perceived", ``vaguely perceived", or ``not perceived", which correspond to the semantic perception scores ($s_i$) of 1.0, 0.5, and 0, respectively. The model's overall capability is then crystallized into the Noteworthy Objects' Perception Recall ($\mathrm{NoPR}$), calculated as the average score across all $N$ targets,
\vspace{-0.1cm}
\begin{equation}
\mathrm{NoPR} = \frac{1}{N}\sum_{i=1}^{N} s_i.
\label{eq:pscore}
\end{equation}
\vspace{-0.1cm}




\noindent\textbf{Knowledge Retention Rate. }To quantify knowledge degradation, we first look to the concept of Backward Transfer (BWT), introduced in prior work on incremental learning~\cite{bwt}. BWT measures the performance change on previous tasks after learning a new one, and is defined as:
\begin{equation}
\mathrm{BWT}=\frac{1}{j-1}\sum_{i=1}^{j-1}\bigl(q_{j,i}-q_{i,i}\bigr),
\end{equation}
where $q_{j,i}$ denotes the performance score on task $i$ after the model has been trained up to task $j$. A negative BWT value signals catastrophic forgetting.

Drawing inspiration from this principle, we propose a metric tailored for the pre-training to fine-tuning paradigm, \textit{i.e.,} the Knowledge Retention Rate (KRR). We conceptualize ``knowledge'' as the model's foundational ability, acquired during pre-training, to perceive and interpret scene elements. A model that truly retains this knowledge should exhibit consistent perceptual capabilities on a held-out set, even after being fine-tuned on a new domain. Therefore, our Noteworthy Objects' Perception Perception Recall ($\mathrm{NoPR}$), when evaluated on the test set designed for measuring forgetting, serves as a direct, operational proxy for this retention. Accordingly, we define the Knowledge Retention Rate (KRR):
\vspace{-0.1cm}
\begin{equation}
\mathrm{KRR} \;=\; \frac{\operatorname{NoPR}(\hat{\Phi};\mathcal{D}_{\mathrm{test}})}
{\operatorname{NoPR}(\Phi_0;\mathcal{D}_{\mathrm{test}})} \,,
\qquad
\label{eq:kr_def}
\end{equation}
where $\Phi_0$ is the original pre-trained model, $\hat{\Phi}$ is the model after fine-tuning, and $\mathcal{D}_{\mathrm{test}}$ is our test set. A KRR value approaching 1.0 indicates good preservation of knowledge from the original model, while a value greater than 1.0 suggests that fine-tuning not only preserves the knowledge of the original pre-trained model but also enhances adaptability to the target domain scenarios.

\subsection{Analysis}
\label{analysis}

\noindent\textbf{How Scene Diversity Shapes Forgetting.}
Our evaluation of publicly available, VLM-centric autonomous driving models reveals that data diversity is a decisive factor in performance and knowledge retention. We assessed a suite of models on the three primary tasks outlined in \cref{sec:metrics}, and the results in \cref{tab:advlm evaluation} are telling. Models fine-tuned on multi-source datasets consistently outperform and exhibit a higher Knowledge Retention Rate (KRR) than those trained on single-source data, irrespective of their foundational VLM. For example, RecogDrive~\cite{recogdrive}, which leverages the most varied data sources, secures the highest KRR across different model sizes. This finding establishes a core principle that \textit{(Guideline-1) diverse training corpora are crucial for mitigating catastrophic forgetting.} This is precisely why we constructed a large-scale training corpus that explicitly prioritizes scene diversity, going so far as to introduce and annotate the most recent WOD-E2E~\cite{wode2e} dataset.

\begin{table}[!t]
\centering
\caption{Results before/after fine-tuning (LoRA) on our benchmark.}
\resizebox{\columnwidth}{!}{
\begin{tabular}{l c c c c c}
\toprule
\textbf{Name} & \textbf{Stage} & \textbf{SD} $\uparrow$ & \textbf{T-QA} $\uparrow$ & \textbf{NoPR} $\uparrow$ & \textbf{KRR} $\uparrow$ \\
\midrule
\multirow{2}{*}{InternVL2-1B}   & Pre  & 57.5 & 20.9 & 35.6 &\multirow{2}{*}{56.2\%} \\
\multirow{2}{*}[12pt]{InternVL2-1B}                                & Post & 54.9 & 23.2 & 20.0 & \multirow{2}{*}[12pt]{56.2\%}   \\ 
\addlinespace[2pt]

\cellcolor{gray!10}\multirow{2}{*}{InternVL2-4B}   & \cellcolor{gray!10}Pre  & \cellcolor{gray!10}61.3 & \cellcolor{gray!10}30   & \cellcolor{gray!10}37.1  & \cellcolor{gray!10}\multirow{2}{*}{43.2\%}\\
\cellcolor{gray!10}\multirow{2}{*}[12pt]{InternVL2-4B}                                & \cellcolor{gray!10}Post & \cellcolor{gray!10}53.1 & \cellcolor{gray!10}23.9 & \cellcolor{gray!10}16.0 & \cellcolor{gray!10}\multirow{2}{*}[12pt]{43.2\%}  \\
\addlinespace[2pt]

\multirow{2}{*}{InternVL3-2B}   & Pre  & 63.8 & 31.5 & 50.8 & \multirow{2}{*}{60.6\%}\\
\multirow{2}{*}[12pt]{InternVL3-2B}                                & Post & 61.8 & 25.1 & 30.8 & \multirow{2}{*}[12pt]{60.6\%} \\
\addlinespace[2pt]

\cellcolor{gray!10}\multirow{2}{*}{Qwen2.5VL-3B}   & \cellcolor{gray!10}Pre  & \cellcolor{gray!10}56.6 & \cellcolor{gray!10}28.7 & \cellcolor{gray!10}36.8 & \cellcolor{gray!10}\multirow{2}{*}{64.6\%}\\
\cellcolor{gray!10}\multirow{2}{*}[12pt]{Qwen2.5VL-3B}         & \cellcolor{gray!10}Post & \cellcolor{gray!10}53.1 & \cellcolor{gray!10}22.3 & \cellcolor{gray!10}23.8 & \cellcolor{gray!10}\multirow{2}{*}[12pt]{64.6\%} \\
\addlinespace[2pt]

\multirow{2}{*}{LLAVA-OV-0.5B}   & Pre & 56.8   & 33.7 & 26.2 & \multirow{2}{*}{71\%}\\
\multirow{2}{*}[12pt]{LLAVA-OV-0.5B}                                & Post & 55.1 & 21   & 18.6 & \multirow{2}{*}[12pt]{71\%} \\
\addlinespace[2pt]

\cellcolor{gray!10}\multirow{2}{*}{LLAVA-OV-7B}    & \cellcolor{gray!10}Pre  & \cellcolor{gray!10}63.7 & \cellcolor{gray!10}35.9 & \cellcolor{gray!10}43.3 & \cellcolor{gray!10}\multirow{2}{*}{57.9\%}\\
\cellcolor{gray!10}\multirow{2}{*}[12pt]{LLAVA-OV-7B}                                & \cellcolor{gray!10}Post & \cellcolor{gray!10}56.7 & \cellcolor{gray!10}47.0 & \cellcolor{gray!10}25.1 & \cellcolor{gray!10}\multirow{2}{*}[12pt]{57.9\%} \\
\bottomrule
\end{tabular}
}
\vspace{-0.5cm}
\label{tab:lora}
\end{table}

\noindent\textbf{Fine Tuning Tradeoffs in Knowledge Retention. }Beyond data-induced forgetting, the adaptation strategy itself is critical to knowledge retention. General-purpose VLMs perform poorly on autonomous driving QA due to a significant domain gap, as shown in \cref{tab:AD model comparison}, which necessitates task-specific fine-tuning. The prevailing method, full fine-tuning, involves updating all model parameters. While this approach yields clear gains on in-domain tasks like Traffic-QA, it concurrently leads to a sharp performance drop in element-level perception, falling even below the original base model. This trade-off, exemplified by the comparison between RecogDrive-8B in \cref{tab:advlm evaluation} and its base VLM InternVL-8B in \cref{tab:AD model comparison}, confirms that \textit{(Guideline-2) aggressive, full-parameter tuning on driving data introduces severe catastrophic forgetting.}

\begin{figure}[!t]
  \centering
    \includegraphics[width=\linewidth]{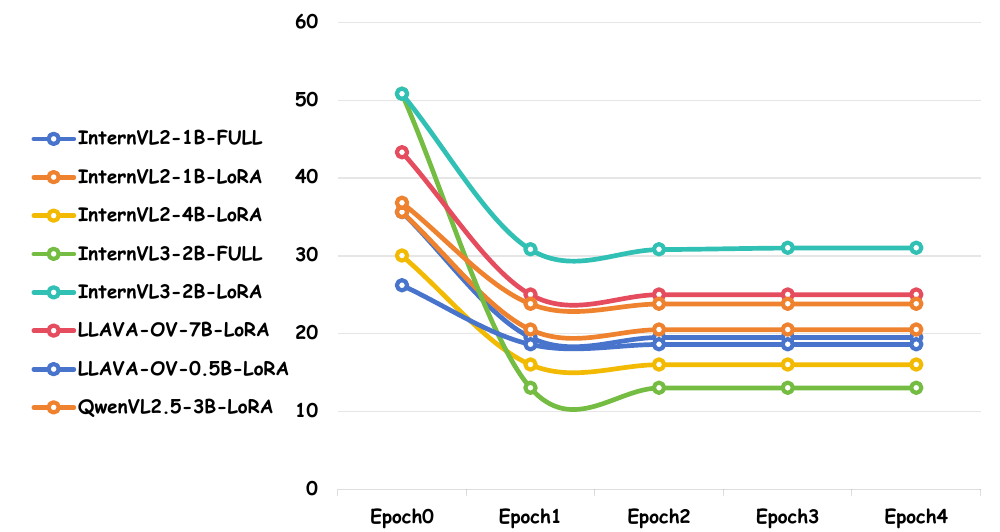}
    \caption{\textbf{Noteworthy Objects’ Perception} Noteworthy Objects’ Perception Recall across fine-tuning epochs on our benchmark. Each curve corresponds to a specific backbone and tuning strategy.}
    \label{fig:elements_perception}
    \vspace{-0.5cm}
\end{figure}

A common strategy to counteract such forgetting is a Parameter-Efficient Fine-Tuning (PEFT) method like LoRA. However, our investigation revealed a counter-intuitive phenomenon. When we applied LoRA fine-tuning to base models using our diverse training corpus, most models' performance on our challenging, long-tail evaluation set did not improve; it actually degraded, as shown \cref{tab:lora}. This suggests that \textit{(Guideline-3) the minimal updates from simple LoRA itself are insufficient to enhance a model's capability in truly rare, long-tail scenarios, despite its better knowledge retention rate than full-parameter tuning (see \cref{fig:elements_perception})}.

\section{Our Approach: Driving Expert Adapter}


Motivated by our analysis and guidelines in \cref{analysis}, we propose the Driving Expert Adapter (DEA) to mitigate the trade-off between knowledge acquisition and retention. DEA avoids destructive parameter modification by injecting expertise through two lightweight components. The Prompt Adapter (PA) encodes driving knowledge into the prompt, a parameter-free strategy to reduce forgetting. Meanwhile, the Task-Adaptive Expert Module (TAEM), a dynamic Mixture-of-Experts, addresses the failure of naive fine-tuning in long-tail scenarios by adaptively handling diverse tasks. This allows the model to gain specialized capabilities while preserving its pretrained knowledge.


\noindent\textbf{Prompt Adapter.}
Our analysis in \cref{analysis} revealed that full fine-tuning leads to catastrophic forgetting, primarily because the model overfits to the specific phrasing of training questions and loses its generalization ability. To address this challenge without destructive parameter updates, we first introduce the Prompt Adapter. This design is inspired by findings that prompt engineering can inject auxiliary knowledge without modifying model weights. However, conventional hand-crafted templates are too rigid for the rich variations of real-world scenes and risk entangling performance with particular wordings. Therefore, our Prompt Adapter utilizes data-driven, learnable and adaptively selected prompts that encode task priors in a more flexible and robust manner.

Operationally, it learns a set of trainable embeddings, each corresponding to a core query family and its associated sub-skills. During inference, the adapter first computes a semantic representation of the input question. It then uses this representation to retrieve the most relevant prompt embeddings, which are prepended to the original input. Crucially, this entire process occurs without modifying the VLM's core parameters. This mechanism stabilizes the model's task interpretation, enhances its generalization to novel phrasings, and thereby effectively mitigates catastrophic forgetting.

\begin{figure}[!t]
  \centering
    \includegraphics[width=\linewidth]{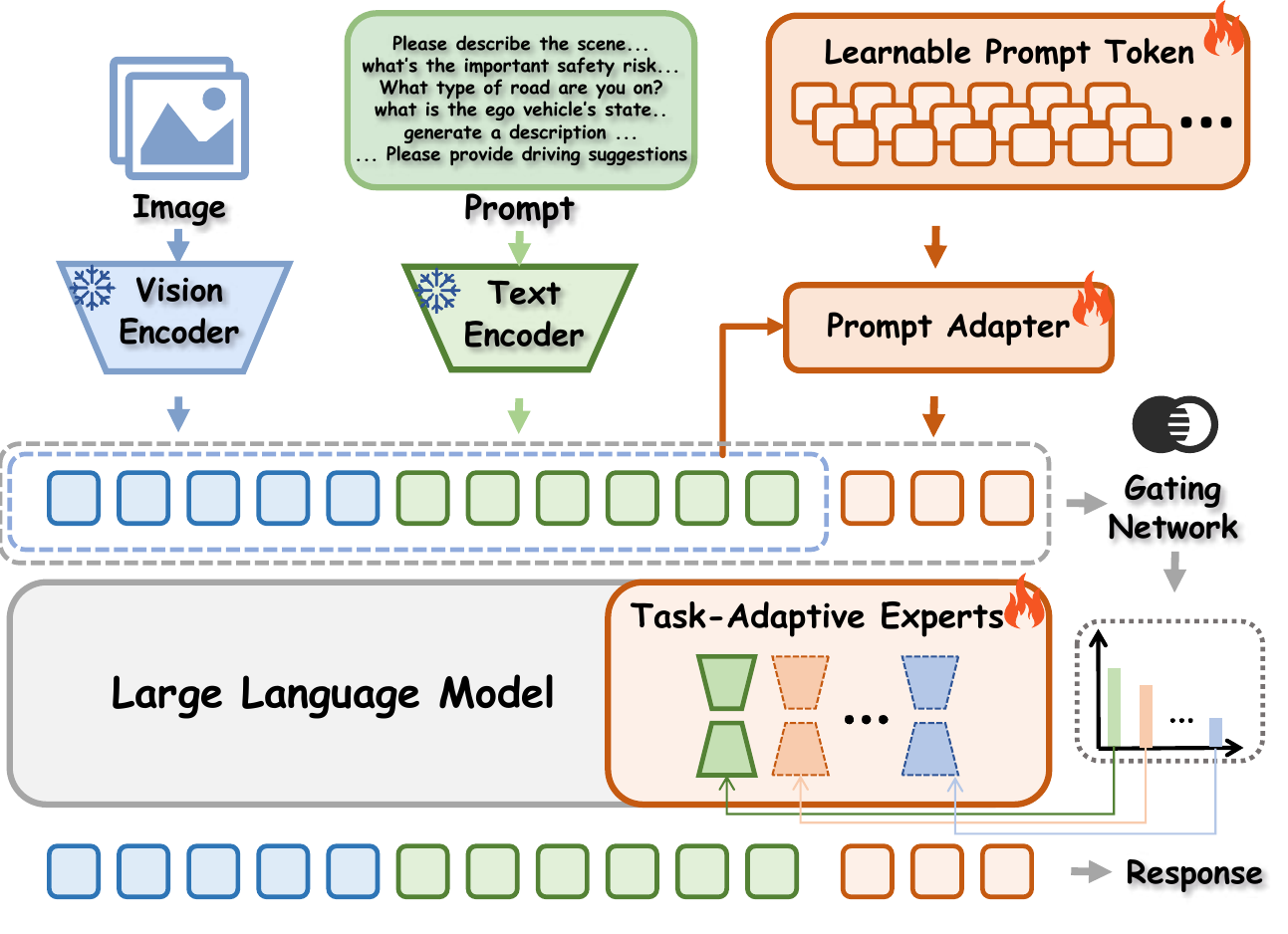}
    \vspace{-0.7cm}
    \caption{\textbf{Illustration of Driving Expert Adapter.} Our framework comprises a Prompt Adapter that selects the most suitable learnable prompt tokens for the current scenario, and a Task-Adaptive Expert Module that leverages a gated network over all tokens to activate the LoRA experts most appropriate for the current scene.}
    
    \label{fig:method_image}
    \vspace{-0.5cm}
\end{figure}

\noindent\textbf{Task-Adaptive Expert Module.}
While the Prompt Adapter strengthens linguistic grounding, it does not solve the second half of our dilemma that the inability of simple, parameter-efficient methods like LoRA to master complex, long-tail scenarios. To bridge this critical capability gap, we designed the Task-Adaptive Expert Module (TAEM). 

Inspired by Mixture-of-Experts (MoE), TAEM operates as an external, modular system coupled with a completely frozen VLM. It comprises multiple ``experts", each tailored to a distinct driving scenario (\textit{e.g.}, navigating dense urban intersections, merging onto highways, or handling adverse weather). A lightweight gating network serves as the module's dynamic router, which processes the enhanced prompt from the Prompt Adapter alongside scene-specific cues (like visibility or traffic density) to select and fuse the outputs of the most relevant experts. This design directly addresses the limitations of monolithic adapters like LoRA. Instead of a single, general-purpose adapter, TAEM provides a powerful and adaptable committee of specialists. This enables the model to inject deep, structured, and context-aware driving knowledge precisely when needed, allowing it to excel in the long-tail scenarios where single-adapter methods fall short.



In synergy, our architecture tackles the core dilemma discussed in \cref{analysis}. The Prompt Adapter ensures stability by compressing knowledge into prompts, preventing the catastrophic forgetting seen in full fine-tuning. The TAEM provides adaptability, addressing the weakness of naive LoRA by deploying specialized experts for long-tail scenarios.

%% file: sec/4_Experiment.tex
\section{Experiments}

We apply our proposed Driving Expert Adapter (DEA) to Qwen2.5VL-3B, which also serves as the foundation model for ImpromptuVLA \cite{impromptuvla}. Experiments and analysis of our approach conducted on additional foundation VLMs are provided in the Appendix due to space limitations.

\subsection{Main Results \& Ablation Study} The main results in \cref{tab:main_results} confirm that our Driving Expert Adapter (DEA) framework successfully resolves the critical trade-off between adaptability and stability. Firstly, Full fine-tuning performs well on T-QA but incurs severe catastrophic forgetting, motivating the use of DEA. On one hand, DEA demonstrates strong adaptability by achieving significant performance gains of 2.2\% on scene description and 12.3\% on Traffic-QA, respectively, when compared to the original, non-fine-tuned Qwen2.5VL-3B baseline. This result refutes the initial observation that LoRA fine-tuning is incapable of enhancing the model's comprehension of long-tail scenarios. On the other hand, and more importantly, DEA effectively mitigates catastrophic forgetting. While any form of fine-tuning introduces some knowledge loss, our method's advantage becomes clear. Notably, ImpromptuVLA-3B, which employs full fine-tuning on the same base model, suffers from severe forgetting. In stark contrast, our DEA maintains a knowledge retention rate of 79.0\%, substantially outperforming ImpromptuVLA-3B's 68.4\%. Even when fine-tuned using the same dataset, the DEA method still achieves a knowledge retention rate of 76.0\%, which further shows that the training set we constructed helps reduce forgetting. This superior knowledge preservation directly validates our central hypothesis. As further substantiated by the results in \cref{tab:main_results}, it is the complementary action of the Prompt Adapter and the Task-Adaptive Expert Module that enables the model to acquire specialized driving skills without the destructive side effects inherent to full-parameter updates.

\begin{table}
\caption{Evaluation results of different fine-tuning strategies on Qwen2.5VL-3B, which is also the VLM in ImpromptuVLA-3B.}
\label{tab:main_results}
\centering
\resizebox{\columnwidth}{!}{
\begin{tabular}{lcccc}
\toprule
\textbf{Method} & \textbf{KRR} $\uparrow$ & \textbf{SD} $\uparrow$ & \textbf{T-QA} $\uparrow$ &  \textbf{NoPR} $\uparrow$   \\
\midrule
\midrule
ImpromptuVLA-3B  & 68.4\% & 59.1 & 33.0 & 25.2  \\
ImpromptuVLA+DEA-3B  & 76.0\% & 58.7 & 38.6 & 27.8  \\
\midrule
Base (Qwen2.5VL-3B) &  - & 56.6& 28.7 & 36.8      \\
Base+Full        & 64.3\% & 54.7 & 35.4 & 23.6     \\
Base+LoRA         & 64.6\% & 53.1 & 22.3 & 23.8 \\
Base+TAEM    &  74.4\% & 57.7 & 35.0   &27.4 \\
Base+TAEM+PA (DEA)  & 79.0\% & 58.8 & 41.0 & 29.0  \\

\bottomrule
\end{tabular}
}
\vspace{-0.5cm}
\end{table}

\subsection{Robustness to the Choice of LLM Judge}
\label{sec:llm_judge_ex}
\begin{figure}
  \centering
    \includegraphics[width=0.95\linewidth, trim={0.1cm 0.2cm 0.2cm 0.1cm}, clip]{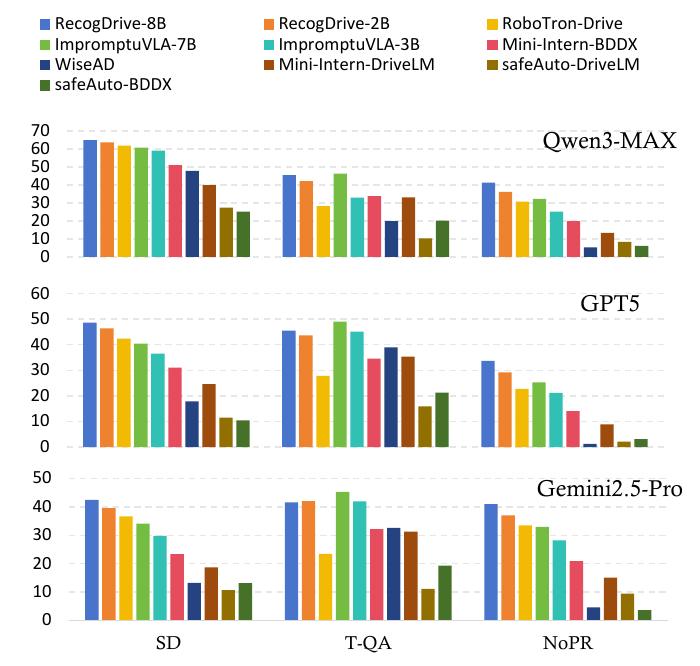} 
    \caption{Comparison among different LLM judges.}
    \label{fig:judge_comparison}
\end{figure}

Fig.~\ref{fig:judge_comparison} validates the robustness of our LLM Judge. We repeated the entire assessment in \cref{tab:advlm evaluation} using a panel of diverse and representative models, including Qwen3-Max~\cite{qwen3}, Gemini-2.5-Pro~\cite{gemini}, and GPT-5~\cite{gpt5}. The results demonstrate remarkable consistency across all settings, the judges preserve nearly identical performance rankings for the competing methods, with only minor fluctuations. This finding confirms that our evaluation primarily captures the intrinsic differences between the autonomous driving systems, rather than being an artifact of a specific LLM judge's bias. 


\begin{figure}
  \centering
    \includegraphics[width=\linewidth]{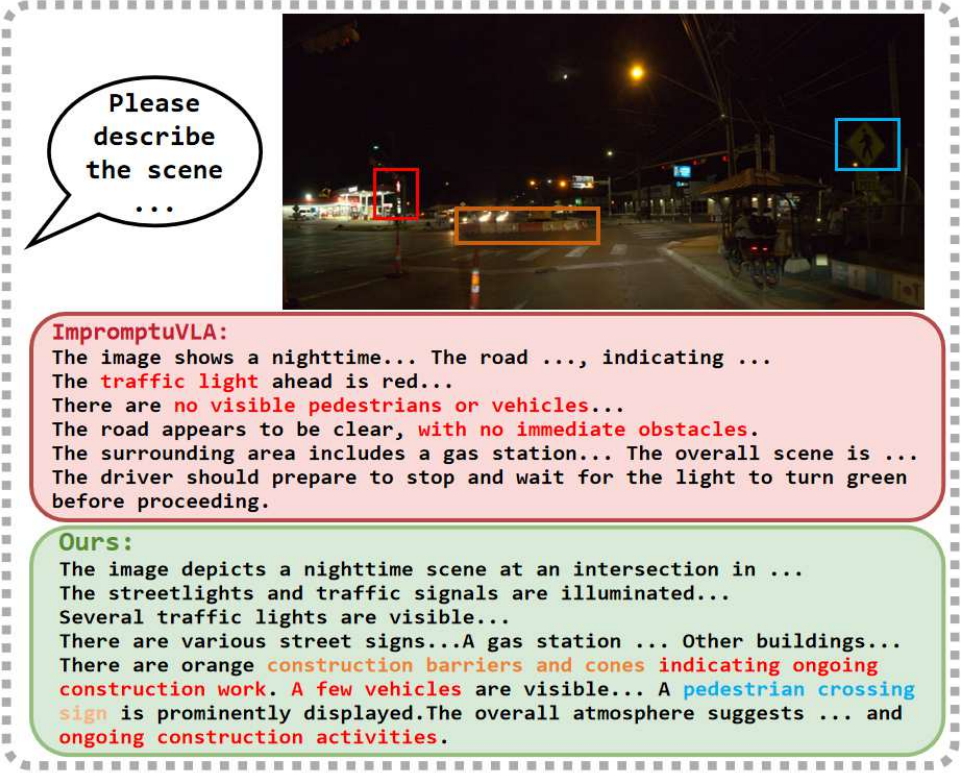}
    \caption{\textbf{Qualitative comparison. } Bounding boxes indicate key objects and corresponding red text provides their descriptions.}
    \label{fig:vis_forgetting}
    \vspace{-0.5cm}
\end{figure}

\subsection{Qualitative Analysis}
In Fig.~\ref{fig:vis_forgetting}, we compare our method with ImpromptuVLA on a nighttime intersection. Our model correctly identifies multiple risk factors, including traffic and a construction zone, which is a typical long-tail scenario easily forgotten during fine-tuning. In contrast, ImpromptuVLA overlooks these critical elements, incorrectly states there are no obstacles, and thus underestimates the potential danger. This comparison demonstrates that our method preserves much richer, fine-grained perception, which is crucial for reliable reasoning in safety-critical environments.

%% file: sec/5_Conclusion.tex
\section{Conclusion}

In this work, we conduct the first systematic study of catastrophic forgetting in VLM-based autonomous driving, a critical yet overlooked challenge. We introduce Fidelity Driving Bench to quantify knowledge degradation, especially in long tail and safety critical scenarios. To address this, we propose the Driving Expert Adapter (DEA), a plug-and-play framework that injects driving specific expertise without destructively overwriting the pretrained model. Experiments demonstrate that our approach substantially mitigates forgetting while maintaining strong driving performance.

\section*{Acknowledgements} This work was supported in part by the Natural Science Foundation of China (Grant No.62503323).